\documentclass[10pt,twocolumn,letterpaper]{article}

\usepackage[pagenumbers]{cvpr} %

\def\paperID{81} 
\def\confName{3DV\xspace}
\def\confYear{2026\xspace}

\usepackage[utf8]{inputenc} 
\usepackage[T1]{fontenc}    
\usepackage{url}            
\usepackage{booktabs}       
\usepackage{amsfonts}       
\usepackage{nicefrac}       
\usepackage{microtype}      
\usepackage[table,xcdraw,dvipsnames]{xcolor}  
\PassOptionsToPackage{dvipsnames, table}{xcolor}
\usepackage{xspace}
\usepackage{cuted}          
\usepackage{amsmath}        
\usepackage{caption}
\usepackage{subcaption}
\usepackage{graphicx}
\usepackage{framed}         
\usepackage{multicol}       
\usepackage{multirow}
\usepackage{makecell}       
\usepackage{pifont}         %

\newcommand{\cmark}{\textcolor{green!60!black}{\ding{51}}}  %
\newcommand{\xmark}{\textcolor{red!70!black}{\ding{55}}}    %

\newcommand{\inlineColorbox}[2]{\begingroup\setlength{\fboxsep}{1pt}\colorbox{#1}{\hspace*{2pt}\vphantom{Ay}#2\hspace*{2pt}}\endgroup}

\definecolor{ModelGreen}{RGB}{213,232,212}
\definecolor{LightGreen}{RGB}{235, 250, 235}  
\definecolor{MediumLightGreen}{rgb}{0.76, 0.93, 0.76}  
\definecolor{MediumLightBlue}{rgb}{0.75, 0.88, 0.98}   
\definecolor{LightBlue}{rgb}{0.88, 0.95, 1.0}   
\definecolor{LightPink}{rgb}{1.0, 0.92, 0.95}   
\definecolor{MediumLightPink}{rgb}{0.98, 0.80, 0.88} 
\definecolor{LightOrange}{rgb}{1.0, 0.94, 0.84}      
\definecolor{MediumLightOrange}{rgb}{0.98, 0.83, 0.63} 
\definecolor{LightGray}{gray}{0.9}

\definecolor{ComplexColor}{RGB}{244, 159, 10}
\definecolor{SimpleColor}{RGB}{156,236,91}

\definecolor{VividGreenModel}{RGB}{179,255,102}
\definecolor{VividBlueModel}{RGB}{102,178,255}

\newcommand{\ourdataset}{CompMo\xspace}
\newcommand{\ourdatasetextended}{Complex Motion Dataset\xspace}
\newcommand{\ourtask}{DMC\xspace}
\newcommand{\ourmethod}{DEMO\xspace}
\newcommand{\ourmethodextended}{Dense Motion Captioning Model\xspace}

\usepackage[most]{tcolorbox} 
\usepackage{xcolor}

\usepackage[subtle]{savetrees} %

\newcommand{\appendixref}[2]{%
    \if\sepappendix1%
    #1%
    \else%
    #2%
    \fi%
}

\def\sepappendix{0}

\definecolor{cvprblue}{rgb}{0.21,0.49,0.74}
\usepackage[pagebackref,breaklinks,colorlinks,citecolor=cvprblue]{hyperref}

\title{Dense Motion Captioning}

\begin{document}

\author{
Shiyao Xu$^{1}$ \quad Benedetta Liberatori$^{1}$ \quad Gül Varol$^{2}$ \quad Paolo Rota$^{1}$ \\
\small{$^{1}$University of Trento \quad $^{2}$LIGM, Ecole des Ponts, IP Paris, Univ Gustave Eiffel, CNRS}\\
\tt\small{shiyao.xu@unitn.it}\\
{\tt\small \href{https://xusy2333.com/demo}{xusy2333.com/demo}}
}

\maketitle

\begin{strip}
\begin{minipage}{\textwidth}\centering
\vspace{-30pt}
\includegraphics[width=\textwidth]{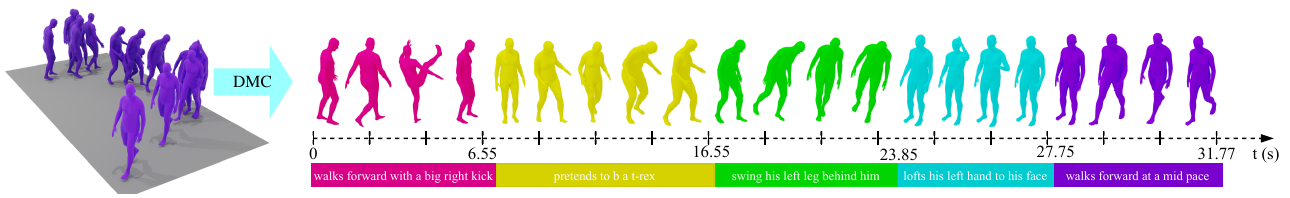}
\captionof{figure}{\textbf{Dense Motion Captioning (\ourtask).} We present \ourtask, a task that localizes and generates detailed segment-level captions with accurate temporal boundaries in 3D human motion sequences. To support this task, we construct \ourdataset, the first large-scale 3D motion-language dataset providing dense captions for multiple temporal segments within each motion sequence. Each sequence contains between 2 and 10 atomic actions, and every action is annotated with precise timestamps and a descriptive caption.}
\label{fig:teaser}
\end{minipage}
\end{strip}

\begin{abstract}
Recent advances in 3D human motion and language integration have primarily focused on text-to-motion generation, leaving the task of motion understanding relatively unexplored.
We introduce Dense Motion Captioning, a novel task that aims to temporally localize and caption actions within 3D human motion sequences. 
Current datasets fall short in providing detailed temporal annotations and predominantly consist of short sequences featuring few actions.
To overcome these limitations, we present the \ourdatasetextended (\ourdataset), the first large-scale dataset featuring richly annotated, complex motion sequences with precise temporal boundaries. 
Built through a carefully designed data generation pipeline, \ourdataset includes \textit{60,000} motion sequences, each composed of multiple actions ranging from at least two to ten, accurately annotated with their temporal extents.
We further present \ourmethod, a model that integrates a large language model with a simple motion adapter, trained to generate dense, temporally grounded captions. 
Our experiments show that \ourmethod substantially outperforms existing methods on \ourdataset as well as on adapted benchmarks, establishing a robust baseline for future research in 3D motion understanding and captioning.
\end{abstract}

\vspace{-10pt}
\section{Introduction}
Recently, there has been a growing interest in integrating 3D human motion and language modalities. 
Most progress in this area has focused on text-to-motion generation~\cite{petrovich2022temos,jiang2024motiongpt,Fg-T2M++, t2mgpt, motionagent, guo2025motionlab}, which involves synthesizing 3D human movements from natural language descriptions, and motion editing~\cite{athanasiou2024motionfix, guo2025motionlab, SALAD_2025_CVPR}, where existing motion sequences are modified according to textual instructions.
These tasks have advanced rapidly, driven by the development of datasets that pair 3D human motions with language descriptions~\cite{KITML,Guo_HumanML3D,BABEL:CVPR:2021,lin2023motionx}.

In contrast, 3D human motion understanding remains in its infancy.
While some recent works have begun to explore this direction, most efforts focus on relatively simple tasks such as motion-to-text retrieval~\cite{petrovich23tmr,tmr++, car_eccv2024} or captioning of short, isolated motion sequences~\cite{chuan2022tm2t,jiang2024motiongpt, zhou2024avatargpt, motionagent}. 
Understanding longer and more complex motion sequences with temporal precision is crucial for applications that require a detailed understanding of human activities. 
For example, by lifting 2D videos into 3D motion representations and generating temporally grounded descriptions from this data, we can develop systems that go beyond traditional video analysis.
This approach allows for a more accurate, body-centric understanding, especially in situations where subtle nuances of motion are crucial.

Motivated by this, we introduce Dense Motion Captioning (\ourtask) as a new task and experimental setting, which involves detecting all semantically meaningful actions in a motion sequence, captioning them, and determining their precise start and end times. 
Unlike traditional single-motion captioning, this task involves parsing a continuous stream of motion and segmenting it into temporally localized action units.

A major limitation of existing benchmarks is their lack of complex motion sequences as well as precise annotations. 
Most available datasets contain only isolated actions or a few simple actions concatenated together, or suffer from noisy annotations, where the descriptions or labels are fragmented and lack consistency. 
In our preliminary experiment on the HumanML3D dataset (see Sec.~\ref{sec:preliminary}), we aim to assess whether current motion captioning models can maintain their performance when handling longer motion sequences containing more than a single action. Our findings indicate a notable performance drop under these conditions.
To address this limitation, we introduce the \ourdatasetextended (\ourdataset), a large-scale dataset specifically designed for dense motion captioning. 
As illustrated in \cref{fig:teaser}, it features extended motion clips with multiple actions. 
Each action is annotated with a detailed caption and temporal boundaries. 
Alongside the dataset, we design \ourmethodextended (\ourmethod), a strong baseline that generates detailed, temporally aligned captions from long and complex 3D motion sequences.
\ourmethod is composed of a Large Language Model (LLM) and a simple motion adapter. It is trained in two stages: first, to align motion and language modalities, and second, to finetune the model for dense caption generation. 
We evaluate it on \ourdataset and existing motion-language datasets repurposed for the \ourtask setting, establishing the first comprehensive benchmark for this task.

In summary, this work makes three main contributions. First, we introduce \ourtask, a novel task which aims to generate sequences of textual descriptions for complex motions, with temporal boundaries. Second, we present \ourdataset, a large-scale dataset specifically curated for this task, featuring rich annotations that capture diverse and intricate human motions across multiple scenarios. Finally, we provide \ourmethod, a strong baseline model along with comprehensive experiments, demonstrating the effectiveness of our approach and fostering future research in this area.

\section{Related Work}
\label{sec:related_work}
\noindent\textbf{3D Human Motion-Language Datasets.} 
Recent years have seen the emergence of datasets designed to advance research in 3D human motion generation and understanding, particularly those that pair motion data with natural language descriptions, with the first effort being the KIT‑ML dataset~\cite{KITML}.
Subsequent efforts~\cite{BABEL:CVPR:2021,Guo_HumanML3D}, significantly scaled the scope of motion-language datasets through crowdsourced annotation of 3D motion clips derived from existing mocap sequences, including AMASS~\cite{AMASS:ICCV:2019} and HumanAct12~\cite{HumanAct12}. 
BABEL~\cite{BABEL:CVPR:2021} annotates motion clips at two abstraction levels: overall sequence categories (\eg, “play basketball”) and subsequence action labels accompanied by durations (\eg, “dribble ball with left hand”, “run”), while many of which contain “transition” in-between.
In HumanML3D~\cite{Guo_HumanML3D}, each motion clip is instead treated as a single semantic unit and described with three natural language sentences from different annotators. 
In contrast, the recent FineMotion~\cite{Wu_FineMotion_Dataset} re‑annotates the same motion sequences in HumanML3D, but segmenting them at uniform temporal intervals, irrespective of action semantics.
Each snippet is labeled with fine-grained body-part movement descriptions (\eg, “raise your hands up to your head”) rather than action-centric labels or descriptions. 
MotionX~\cite{lin2023motionx} and its successor MotionX++~\cite{zhang2025motionx++} shift the emphasis from more detailed captions toward enriching modalities. 
MotionX uses SMPL-X whole body pose annotations, covering body, hands, and facial expressions, paired with semantic labels.
MotionX++ goes further by adding synchronized RGB video and audio data alongside pose annotations and textual descriptions. 
We propose \ourdataset, focusing on dense 3D human motion captioning. 
Rather than short labels like those in BABEL, coarse whole-clip captions like in HumanML3D, or snippet-level body-part descriptions as in FineMotion, our dataset provides rich sequence-level natural language descriptions, each annotated with precise temporal timestamps. 
\ourdataset thus establishes a new benchmark for dense motion captioning and motion-language alignment in 3D human motion, an area not yet addressed by existing datasets.

\noindent\textbf{Dense Video Captioning.} 
Dense Video Captioning (DVC) extends standard video captioning by identifying multiple temporal segments in an untrimmed video and generating corresponding textual descriptions for each segment~\cite{krishna2017dense}.
Earlier methods typically followed a two-stage, detect-then-describe paradigm~\cite{Wang_2018_CVPR,krishna2017dense}, whereas recent approaches have shifted towards end-to-end training for improved efficiency and performance~\cite{yang2023vid2seq,chen2021towards,Zhang2022UnifyingED}.
Effective DVC requires both accurate temporal localization and semantic correctness, and evaluation metrics must account for both aspects. 
To address this, DVC evaluation typically combines standard captioning metrics~\cite{vedantam2015cider,banerjee2005meteor} with Intersection over Union (IoU) thresholds.
More recently, SODA~\cite{fujita2020soda} has been introduced as a comprehensive metric that temporally aligns predicted and reference captions before computing METEOR-based scores that penalize redundancy and poor alignment.
We propose Dense Motion Captioning (\ourtask), bringing this paradigm to the domain of 3D human motion understanding, challenging models to generate temporally precise descriptions of human motion.

\begin{figure*}
    \centering
    \includegraphics[width=\linewidth]{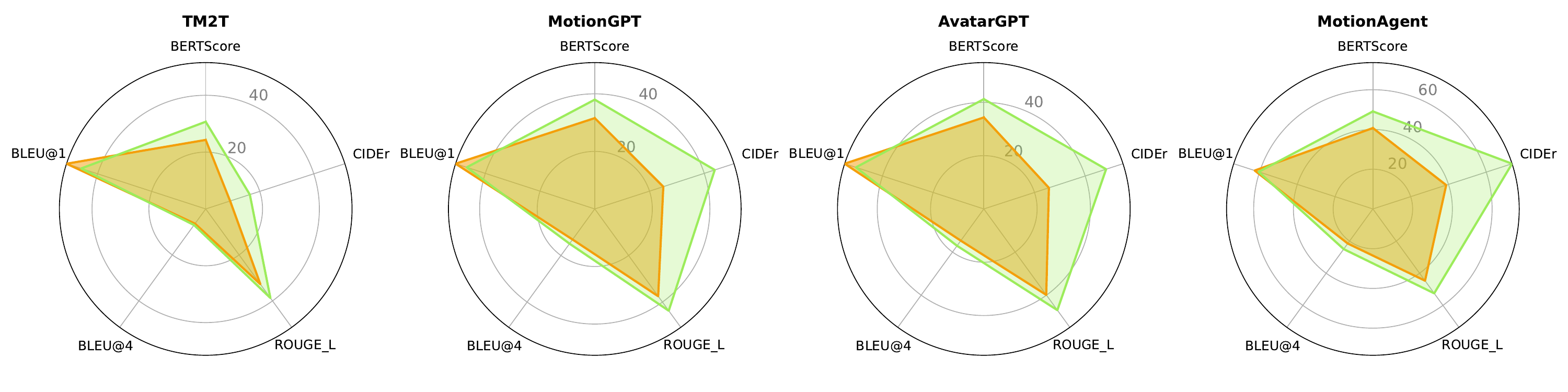}
    \caption{\textbf{Single Motion Captioning performance divided by \emph{\textcolor{SimpleColor}{simple}} and \emph{\textcolor{ComplexColor}{complex}} motion sequences.}
     We report the single motion captioning performance of state-of-the-art motion-language models on the simple and complex subsets of HumanML3D~\cite{Guo_HumanML3D}'s test set, as defined in Sec.~\ref{sec:preliminary}.
     }
    \label{fig:simple_complex}
\end{figure*}

\noindent\textbf{Human Motion Understanding.}
Much of prior work in human motion research has focused on motion generation~\cite{MDM,petrovich2022temos,shafir2024human,huang2024como,petrovich2021action}, \ie, synthesizing realistic 3D human movements from text or other modalities.
More recently, the motion-to-text task has also gained attention, with methods developing unified motion-language models capable of both generating motion from text and describing input motion~\cite{chuan2022tm2t, zhou2024avatargpt, sun2024coma, huang2024como, zhang2024kinmo, li2024human, chen2024motionllm, motionagent}. 
While these demonstrate impressive versatility, their accuracy in motion understanding remains limited, particularly in tasks requiring temporal precision. 
This limitation arises because they are not trained to capture or describe sub-sequences within longer, continuous motions, which is essential for detailed temporal comprehension.

Beyond this, some works explore related but distinct challenges. 
BABEL-TAL~\cite{locate:2024_babeltal} tackles 3D temporal action localization, which involves recognizing actions performed in a 3D motion sequence and precisely identifying their start and end times, albeit with a fixed set of action class labels. 
Similarly, TMR~\cite{petrovich23tmr} shows the use case of moment retrieval by temporally localizing BABEL actions within long sequences. This idea is later extended by UniMotion~\cite{li2024unimotion} to frame-level motion captioning as an initial exploration of dense action recognition. 
However, UniMotion~\cite{li2024unimotion} treats captioning as a retrieval problem with a closed vocabulary of action labels, and they do not provide a quantitative benchmark. In contrast, our method generates free-form descriptions and outputs segment timestamps instead of assigning an action label per frame.

\section{From Simple to Complex Motions}
In this section, we first motivate our study with a preliminary analysis of the widely used HumanML3D dataset~\cite{Guo_HumanML3D} (Sec.~\ref{sec:preliminary}).
We then describe the generation pipeline of our dataset (Sec.~\ref{sec:data}).

\subsection{Can Current Models Understand Complex Human Motions?}\label{sec:preliminary}

The HumanML3D~\cite{Guo_HumanML3D} dataset is widely used to evaluate human motion understanding models thanks to its diverse range of motion sequences of varying complexity.
In this study, we investigate whether the complexity of a motion, specifically, the presence of multiple sub-actions, correlates with the performance of state-of-the-art motion-language models. 
To this end, we partition the mirrored augmented dataset with 29,228 motions into two disjoint subsets: \emph{simple} and \emph{complex} motions.
This partitioning is based on the number of verbs/adverbs in the ground-truth textual descriptions, under the assumption that each verb typically corresponds to a distinct sub-action (\eg, ``a person \textit{sits} down and \emph{crosses} their leg, before \emph{getting up}''). 
Motions described with no more than 1 verb are considered simple, while those with 2 or more are labeled complex. 
This results in 17,512 complex and 11,716 simple motion instances, of which 2,663 and 1,721 are from the test set, respectively. 
We evaluate the performance of several recent models for standard single motion captioning, \ie, generating one description without timestamps,~\cite{chuan2022tm2t,zhou2024avatargpt,motionagent,jiang2024motiongpt} on both subsets.\footnote{We exclude models that have not released code at the time of writing.} 
Fig.~\ref{fig:simple_complex} reports the obtained results in terms of single motion captioning metrics~\cite{chuan2022tm2t}. 
In the vast majority of cases, we observe a considerable drop in performance on the complex subset, highlighting that current state-of-the-art models tend to perform better on simpler samples but struggle to accurately understand and describe longer sequences with multiple sub-actions. 
\textbf{This finding motivates our study}, emphasizing the need for datasets that present greater temporal complexity to better train and evaluate motion-language models, ultimately enabling more precise temporal motion understanding.

\subsection{\ourdataset: A Complex Motion Dataset}
\label{sec:data}

\begin{table}[]
  \centering
  \resizebox{\linewidth}{!}{
  \begin{tabular}{lcccclc}
    \toprule
    Dataset & \makecell{Dataset \\ Size} & \makecell{Avg. \\ Duration (s)} & \makecell{Annotation \\ Type} & Timestamps \\
    \midrule
    KIT-ML~\cite{KITML}         & 3,911  & 10.33 & Sentence & \xmark \\
    HumanML3D~\cite{Guo_HumanML3D} & 14,616 & \multirow{2}{*}{7.1} & \multirow{2}{*}{Sentence} & \multirow{2}{*}{\xmark} \\
    (mirror) & 29,228 & & & \\
    BABEL~\cite{BABEL:CVPR:2021} & 13,220 & 12.26 & Labels & \cmark \\
    MotionX~\cite{lin2023motionx} & 81,084 & 6.4 & Sentence & \xmark \\
    MotionX++~\cite{zhang2025motionx++} & 120,462 & 5.4 & Sentence & \xmark \\
    FineMotion~\cite{Wu_FineMotion_Dataset} & 14,616 & 7.1 & Fine Descriptions & \cmark \\
    \midrule
    \ourdataset (ours) & 60,000 & 39.88 & Dense Captions & \cmark \\
    \bottomrule
  \end{tabular}
  }
    \caption{\textbf{Overview of \ourdataset and prior 3D motion-language datasets.} While existing datasets vary in size, annotation type, and temporal richness, our \ourdataset is the first large-scale dataset designed for DMC with accurate timestamps, enabling more comprehensive modeling of temporally complex motions.}
  \label{tab:dataset_comp}
\end{table}

To address the limitations current models face in handling temporally complex motions, we introduce the \ourdatasetextended (\ourdataset), a new large-scale dataset specifically designed to challenge and advance motion-language models. 
\ourdataset is the first dataset explicitly created for 3D dense motion captioning with precise timestamps, enabling more effective training and evaluation of models.
It features longer motion sequences, providing more temporally extended contexts for dense captioning. 
On average, each motion in \ourdataset\ is annotated with 37.74 words, compared to 12 and 11.06 words in HumanML3D~\cite{Guo_HumanML3D} and BABEL~\cite{BABEL:CVPR:2021}. 
Compared to existing temporally annotated motion datasets, \ourdataset\ represents a significant increase in both scale and complexity (see Tab.~\ref{tab:dataset_comp}).
To support these design goals, we developed a multi-stage pipeline for dataset construction, which we describe in detail below.

\noindent\textbf{Atomic Actions Collection.}
To build a diverse and high-quality dataset for dense motion captioning, we begin by collecting simple human motions paired with textual descriptions. 
We use HumanML3D~\cite{Guo_HumanML3D} as our primary source, as it provides an extensive collection of motion-text pairs encompassing a wide range of human motions, including everyday activities, sports, and artistic movements. 
Following our preliminary analysis (Sec.~\ref{sec:preliminary}), we employ the \textit{simple} set, treating each element as an atomic action aligned with its corresponding atomic description.  

To obtain better alignment between motion and text, we propose two strategies for data collection: \textbf{i)} generated from scratch, and \textbf{ii)} drawn directly from the \textit{simple} set. For the data in \textbf{i)}, we use the diffusion-based MDM-SMPL model proposed in STMC~\cite{petrovich24stmc} to generate the motions from their textual descriptions; Then we use TMR~\cite{petrovich23tmr}, a model that encodes motions and languages into a shared embedding space, as encoder, to calculate the cosine \textit{TMR Similarity} across different modalities, and filter out candidates with low motion-text alignment. To address motion types that are poorly generated, we supplement the dataset with samples from the \textit{simple} HumanML3D set.
The final atomic actions, accompanied by descriptions, contain 7,503 generated from scratch and 3,619 drawn from HumanML3D.

\noindent\textbf{Textual Descriptions Composition.}
Starting from atomic actions, we perform a temporal composition for atomic descriptions by randomly sampling 2 to 10 atomics and combining these into coherent sequences. Each sequence is annotated with precise timestamps, formatted as ``\texttt{<mm:ss:ms: atomic textual description>}''.
To ensure realistic and varied durations, we condition the length of each motion segment on its ground-truth duration from HumanML3D, applying small random perturbations to introduce variability while preserving temporal plausibility. 

\noindent\textbf{Motion Sequences Generation.}
We then generate human motion sequences corresponding to the constructed textual descriptions. 
Inspired by STMC~\cite{petrovich24stmc}, which applies a test-time denoising approach for spatio-temporal motion composition, we also employ the temporal stitching technique of DiffCollage~\cite{zhange2023diffcollage} as well as the body part stitching in combination with MDM-SMPL provided by~\cite{petrovich24stmc}. At each denoising step, we start from the textual description, denoise, stitch the resulting conditions together both temporally and across the relevant body parts, and finally generate the composed motion sequences.

\noindent\textbf{Final Dataset Description.} 
The resulting \ourdataset{} dataset contains 60,000 motion-text pairs with timestamp annotations. 
On average, motion sequences last 39.88 seconds, significantly longer than sequences in existing datasets, reflecting the increased temporal complexity of \ourdataset.
We partition the dataset into training, validation, and test sets, corresponding to the 80\%/10\%/10\% of the data, respectively. 
Additional details on the generation pipeline are provided in Sec.~\ref{subsec:app_data} of the Appendix.

\section{DEMO: Dense Motion Captioning Model}

\begin{figure*}[t]
  \centering
  \includegraphics[width=\linewidth]{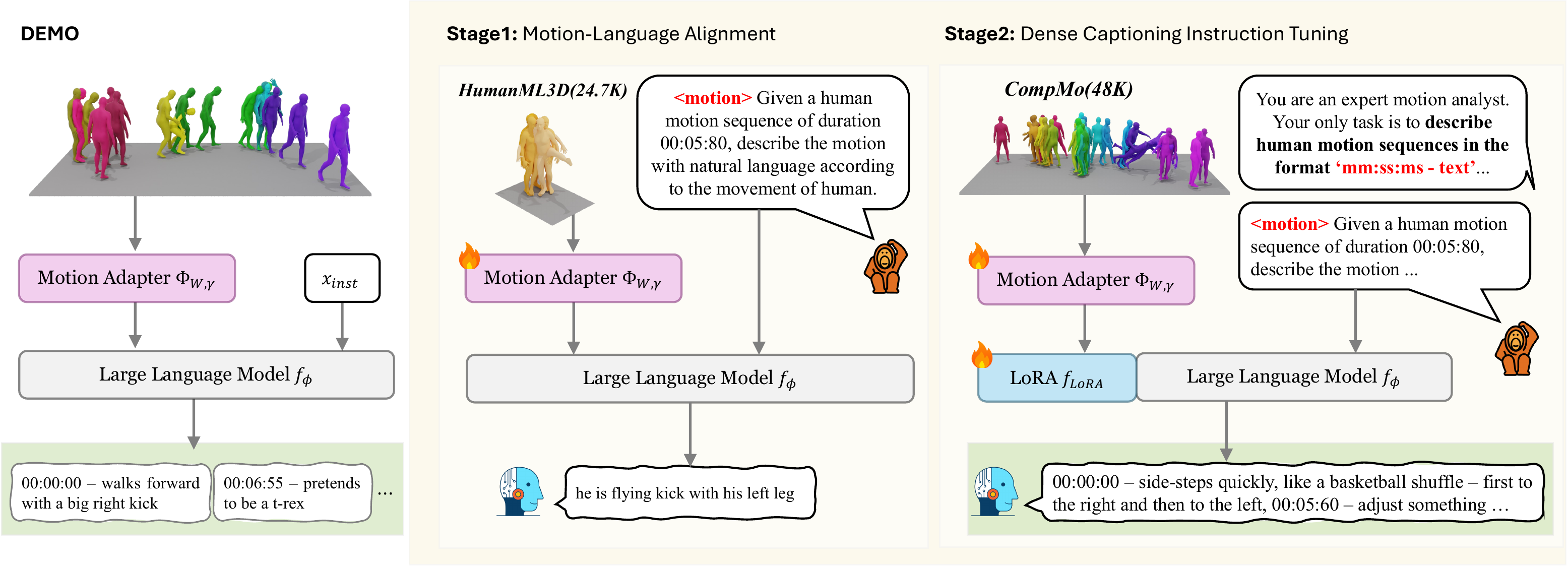} %
  \caption{\textbf{\ourmethod overview :} 
  Given a motion sequence $m$, our method encodes it with the motion adapter $\Phi_{\textbf{W},\gamma}$, which maps it into the language embedding space of the LLM $f_\phi$. Using the resulting motion embeddings and a textual instruction $x_{inst}$, the model generates dense captions with temporal boundaries. Training is conducted in two stages. Here, \raisebox{-0.5mm}{\includegraphics[height=3.5mm]{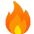}} denotes the subset of parameters being trained.
  }
  \label{fig:method}
\end{figure*}

In this section, we first formalize the dense motion captioning task (\cref{sec:formulation}) then detail our proposed architecture (\cref{sec:method} and \cref{sec:repre}) and training procedure (\cref{sec:training}).  

\subsection{Problem Formulation}\label{sec:formulation}

Given a 3D human motion sequence $m\in\mathbb{R}^{N\times D}$, where $N$ is the number of poses and $D$ is the dimensionality of each pose, Dense Motion Captioning (DMC) consists in generating a sequence $\{(t_i,c_i)\}_{i=1}^M$, where $t_i = (s_i, e_i)\in \mathbb{R}^2$ represents the start and end times of the i-th motion segment, $c_i$ is a caption describing the human motion within that segment, and $M$ is the number of atomic actions detected. 
We define the pose dimensionality as $D = J \times 3$, where $J$ is the number of 3D joints used to represent each pose.
Unlike the traditional single motion captioning task, DMC requires both accurate temporal localization of atomic motion segments and natural language generation.

\subsection{Method}\label{sec:method}

Our architecture, \ourmethod, leverages an LLM, finetuned to autoregressively generate dense, temporally aligned captions from long and complex 3D motion sequences, as illustrated in \cref{fig:method} (left). Let $f_\phi$ denote the LLM, parametrized by $\phi$. Since $f_\phi$ is originally pretrained only on text and vision modalities, it cannot directly process motion data. To address this, we first convert the continuous motion sequence $m\in \mathbb{R}^{N\times D}$ into a language-compatible embedding space that can be processed by $f_\phi$, and then use $f_\phi$ to generate the dense motion descriptions.

\subsection{Motion Representation}
\label{sec:repre}

Prior LLM-based approaches represent a continuous motion by learning a mapping to discrete tokens, \eg, training a vector quantized variational autoencoder (VQ-VAE) to construct a \emph{motion vocabulary}. 
However, this approach suffers from two key limitations: (i) inherent information loss caused by the limited discrete vocabulary~\cite{meng2024rethinking, MDM}, and (ii) the need for an additional, separate training stage for the VQ-VAE.
In contrast, \ourmethod learns a simple continuous mapping from motion to language space using a single network. 
Specifically, a lightweight motion encoder $\gamma$ extracts motion features, which are then adapted into the language domain via a linear projection $\textbf{W}$, eliminating intermediate discretization.

Since the motion sequences in \ourdataset{} can last up to 10 times the duration of those in HumanML3D, this necessitates a scalable and efficient strategy for encoding long sequences. 
Processing the entire motion at once is computationally expensive and often unnecessary, as generating detailed descriptions for short motion segments typically depends only on their immediate temporal context rather than the full sequence.

To address this, we partition the input motion sequence into a series of fixed-size, overlapping windows $\{m^{(i)} \in \mathbb{R}^{W \times D}\}_{i=1}^K$, extracted with a stride $S < W$. Each window $m^{(i)}$ corresponds to a sub-sequence of the full motion $m$ and is processed independently to capture temporally localized motion patterns. The window is first flattened, added with positional embeddings, and then passed through the motion adapter defined as:

\begin{equation}
   \Phi_{\gamma, \mathbf{W}}(m^{(i)}) = \mathbf{W} \cdot \gamma(m^{(i)}),
\end{equation}
where the adapter projects the motion features into the language embedding space of $f_\phi$.

\subsection{Training Strategy}\label{sec:training}

We train \ourmethod to autoregressively generate motion captions given a 3D motion sequence and a textual instruction. 
Given an input motion sequence $m$ and instruction prompt $x_{inst}$ as input, the generation process is modeled as:
\begin{equation}
\begin{aligned}
    p(\mathbf{y} \mid m, x_{inst}) = \prod^{L}_{i=1} p_\theta (y_i | m,  x_{inst}, y_{<i}), %
\end{aligned}
\end{equation}
where $\mathbf{y} = \{y_1, \dots, y_L\}$ is the output caption of length $L$, $p(\cdot)$ is the model's probability distribution over tokens, and $y_{<i}$ denotes the previously generated tokens up to position $i-1$.
The parameter set $\theta$ includes all trainable components of the model.
During training, we optimize the model by maximizing the log-likelihood of the target caption, using the cross-entropy loss:

\begin{equation}
    \mathcal{L} = - \sum_{i-1}^L \log p_\theta (y_i \mid m, x_{inst}, y_{<i}).
    \label{eq:objective}
\end{equation}

As illustrated in \cref{fig:method}, the motion adapter $\Phi_{\gamma, \mathbf{W}}$ and the LLM $f_\phi$ are trained in a two-stage process: first, a motion-language alignment stage to align motion features with the language model’s embedding space, followed by a dense caption instruction tuning stage to enable precise and temporally grounded caption generation. 
While the training objective remains the same as in Eq.~\eqref{eq:objective} in both stages, what differs are the instruction prompts $x_{inst}$, target outputs $\mathbf{y}$, input motion data $m$, and the subsets of parameters in $\theta$ optimized during training.
These stages are described in detail below.

\noindent\textbf{Stage 1: Pretraining for Motion-Language Alignment.}
In this stage, we focus on aligning the motion modality with the language space by training only the motion adapter, \ie, $\theta=\Phi_{\gamma, \mathbf{W}}$ on paired motion-text data.
To achieve this alignment, we use the HumanML3D~\cite{Guo_HumanML3D}, where each motion $m$ consists of a single motion sequence, and $\mathbf{y}$ is the paired ground truth annotation, without timestamps.
The instruction prompt $x_{inst}$ is designed as shown in \cref{fig:method} (center), providing only the overall motion duration. 

\noindent\textbf{Stage 2: Dense Captioning Instruction Tuning.}
In this stage, we instruct the model to generate temporally grounded captions, explicitly including action boundaries and their corresponding timestamps.
We use \ourdataset, where each motion $m$ is a longer, complex sequence, and the target output $\mathbf{y}$ is a sequence of captions paired with their annotated temporal intervals.
The instruction prompt $x_{inst}$ is adapted accordingly to guide the model in producing temporally localized descriptions, as illustrated in \cref{fig:method} (right).
To enable efficient finetuning, we apply LoRA~\cite{hu2022lora} to the language model $f_\phi$, while jointly finetuning the pretrained motion adapter along with LoRA.
Thus, the set of trainable parameters in this stage is $\theta=\{\Phi_{\gamma, \mathbf{W}}, f_{LoRA}\}$. 
This stage equips the model with the ability to generate fine-grained, time-aware descriptions of complex motions.

\begin{table*}%
  \resizebox{\linewidth}{!}{
  \centering
  \begin{tabular}{ll|ccccccc|cc|cc}
    \midrule
    \multirow{2}{*}{\textbf{Method}} & \multirow{2}{*}{\textbf{Dataset}} &
    \multicolumn{7}{c|}{\textbf{Dense Captioning $\uparrow$}} & \multicolumn{2}{c|}{\textbf{Localization $\uparrow$}} & \multicolumn{2}{c}{\textbf{T-M Similarity $\uparrow$}} \\
    & & \textbf{SODA} & \textbf{SODA(B)} & \textbf{CIDEr} & \textbf{METEOR} &  \textbf{ROUGE\_L} & \textbf{BLEU@1} & \textbf{BLEU@4} & \textbf{tIoU \%} & \textbf{F1 \%} & \textbf{TMR} & \textbf{CAR} \\
    \midrule
    UniMotion~\cite{li2024unimotion} & \cellcolor{MediumLightOrange}CompMo & 0.6099 & 12.8090 & 1.0082 & 0.4266 & 0.8479 & 0.7793 & 0.0000 & 36.14 & 4.00 & 0.4930 & 0.3487  \\
    \ourmethod &
    \cellcolor{MediumLightOrange}CompMo & \cellcolor{LightOrange}17.8473& \cellcolor{LightOrange}64.4003& \cellcolor{LightOrange}134.4424& \cellcolor{LightOrange}16.4085& \cellcolor{LightOrange}24.0469& \cellcolor{LightOrange}23.8980& \cellcolor{LightOrange}11.0024& \cellcolor{LightOrange}77.94& \cellcolor{LightOrange}58.21& \cellcolor{LightOrange}0.6832&
    \cellcolor{LightOrange}0.8027%
    \\
    \midrule
    UniMotion~\cite{li2024unimotion} & \cellcolor{MediumLightPink}H3D $\cap$ BABEL & 5.7141 & \cellcolor{LightPink}30.4658 & 6.7170 & 5.0826 & 5.8060 & 5.1651 & \cellcolor{LightPink}0.4375 & 49.95 & \cellcolor{LightPink}22.23 & \cellcolor{LightPink}0.6428 & \cellcolor{LightPink}0.8473 \\
    \ourmethod & \cellcolor{MediumLightPink}H3D $\cap$ BABEL & \cellcolor{LightPink}7.9194 & 25.9654 & \cellcolor{LightPink}7.8090 & \cellcolor{LightPink}5.7625 & \cellcolor{LightPink}6.2919 &\cellcolor{LightPink}5.6936 & 0.1318 & \cellcolor{LightPink}51.56 & 16.40 & 0.6052 & 0.8204\\ 
    \bottomrule
  \end{tabular}
  }
\caption{\textbf{Comparison on Dense Motion Captioning}. We compare the performance of \ourmethod on the proposed \inlineColorbox{MediumLightOrange}{\ourdataset} and on \inlineColorbox{MediumLightPink}{H3D $\cap$ BABEL}. We measure dense captioning, temporal localization, and motion-caption alignment accuracy. Best results are highlighted.}
  \label{tab:main_comp_final}
\end{table*}

\section{Experiments}

\noindent\textbf{Datasets and Settings. }
We conduct DMC experiments on two datasets: our proposed \textbf{\ourdataset}, and the intersection of HumanML3D~\cite{Guo_HumanML3D} and BABEL~\cite{BABEL:CVPR:2021}, following the setup introduced in UniMotion~\cite{li2024unimotion}.
CompMo comprises 60,000 motion sequences paired with dense captions, divided into 48,000/12,000 for training/testing. 
The dataset adopted from~\cite{li2024unimotion}, here denoted with \textbf{H3D $\cap$ BABEL}, is constructed from the overlapping subset of HumanML3D and BABEL, and consists of 7,056/1,325 motion sequences paired with frame-level annotations for training/testing. 
Additionally, we use \textbf{HumanML3D} for the first stage of our training. We adopt the train+val split of the mirrored augmented dataset, including 23384+1460 motion sequences, each annotated with three descriptions. 
During training, we randomly sample one of the associated annotations at each step.

\noindent\textbf{Metrics.}
We quantify DMC performance using dense captioning accuracy, temporal localization accuracy, and motion-caption alignment.
For dense captioning, we follow dense video captioning literature~\cite{ventura2025chapter, yang2023vid2seq, krishna2017dense, chuan2022tm2t}, computing captioning metrics: CIDEr~\cite{vedantam2015cider}, METEOR~\cite{banerjee2005meteor}, ROUGE\_L~\cite{lin-2004-rouge}, BLEU~\cite{papineni-etal-2002-bleu}, over matched prediction-reference pairs within the IoU thresholds of $\{0.3, 0.5, 0.7, 0.9\}$, reporting the average results on the matched pairs. 
We also use SODA~\cite{fujita2020soda} with two different linguistic metrics, METEOR~\cite{banerjee2005meteor} and BertScore~\cite{zhang2019bertscore} (corresponding to SODA and SODA(B) in Tab.~\ref{tab:main_comp_final}), for overall caption evaluation.
For temporal localization, we follow~\cite{ventura2025chapter}, using a greedy algorithm to select the best matching with the highest IoU, then computing the mean IoU for all matched pairs to get the overall tIoU and F1 score.  
For motion-caption alignment, following prior work on image and video captioning~\cite{zhang2022image_caption, krishna2017dense}, we measure the cross-modal distance between motion sequences and their generated captions. Specifically, we calculate the cosine similarity between motions and texts in the joint embedding space of TMR~\cite{petrovich23tmr}. To further assess the sequential alignment, we adopt the CAR~\cite{car_eccv2024} score, a recent work that improves the motion-text retrieval by introducing negative samples generated through event-sequence shuffling, encouraging the model to achieve better temporal alignment,
where we retrieve motions given a set of shuffled and generated event sequence captions from the test set with 32 samples.

\noindent\textbf{Implementation Details. }
We use 3D joint representations with $J = 22$ joints. We set the window size and stride to $W=16$, $S=8$. Our $f_\phi$ is initialized with LLaMA-3.1-8B-Instruct~\cite{llama3.1}, while $\gamma$ is an MLP. 
Training takes approximately 3.5 hours on 2 NVIDIA RTX 6000 Ada GPUs.
Additional implementation details are provided in Sec.~\ref{subsec:app_exp} of the Appendix.

\begin{table*}
  \centering
  \resizebox{\linewidth}{!}{
  \begin{tabular}{l|ccccccc|cc|cc}
    \toprule
    \multirow{2}{*}{\textbf{Method}}&
    \multicolumn{7}{c|}{\textbf{Dense Captioning $\uparrow$}} & \multicolumn{2}{c|}{\textbf{Localization $\uparrow$}} & \multicolumn{2}{c}{\textbf{T-M Similarity $\uparrow$}} \\
     & \textbf{SODA} & \textbf{SODA(B)} & \textbf{CIDEr} & \textbf{METEOR} &  \textbf{ROUGE\_L} & \textbf{BLEU@1} & \textbf{BLEU@4} & \textbf{tIoU \%} & \textbf{F1 \%} & \textbf{TMR} & \textbf{CAR} \\
    \midrule
    \rowcolor{LightGreen}\multicolumn{12}{l}{\textit{\textbf{Dataset Generation}}} \\
    Concat GT & 1.9910 & 41.5498 & 8.2427 & 1.9572 & 4.0158 & 4.2401 & 0.0428 & 61.45 & 27.52  & 0.5414 & 0.4505\\
    Smooth GT & 1.9561 & 41.5586 & 8.1089 & 1.8835 & 3.9398 & 4.1223 & 0.0230 & 61.08 & 26.74 & 0.5306 & 0.4977 \\
    Denoise only from random & 12.1643 & 62.4457 & 80.9095 & 11.9174 & 18.2653 & 18.3024 & 5.1632 & 77.92 & 57.32 & 0.5680 & 0.7895 \\
    Denoise only from GT & 13.3860 & 55.2276 & 94.7040 & 12.7457 & 17.5265 & 17.7187 & 7.6551 & 69.89 & 43.00 & 0.5754 & 0.7987\\
    \rowcolor{gray!10}
    CompMo & 17.8473 & 64.4003 & 134.4424 & 16.4085 & 24.0469 & 23.8980 & 11.0024 & 77.94 & 58.21 & 0.6832 & 0.8027\\
    \midrule
    \rowcolor{LightOrange}\multicolumn{12}{l}{\textit{\textbf{Training Stages}}} \\
    Stage 2 & 1.6521 & 28.4648 & 4.5059 & 1.2444 & 2.0972 & 2.3754 &  0.0362 & 49.45 & 14.28 & 0.6056 & 0.5987\\
    \rowcolor{gray!10}
    Stage 1+2 & 17.8473 & 64.4003 & 134.4424 & 16.4085 & 24.0469 & 23.8980 & 11.0024 & 77.94 & 58.21 & 0.6832 & 0.8027\\
    \midrule
    \rowcolor{LightPink}\multicolumn{12}{l}{\textit{\textbf{Motion Representation}}} \\
    VQ-VAE & 2.3398 & 43.3563 & 7.6868 & 2.0440 & 3.4973 & 3.6243 & 0.0778 & 60.76 & 26.60 & 0.5881 & 0.6282 \\ 
    \rowcolor{gray!10}
    $\Phi_{\textbf{W},\gamma}$  & 17.8473 & 64.4003 & 134.4424 & 16.4085 & 24.0469 & 23.8980 & 11.0024 & 77.94 & 58.21 & 0.6832 & 0.8027\\
    \bottomrule
  \end{tabular}}
  \caption{\textbf{Ablation Study.} We assess the contribution of different components by ablating variations in \inlineColorbox{LightGreen}{dataset generation} (data-level), as well as \inlineColorbox{LightOrange}{training stages} and \inlineColorbox{LightPink}{motion representation} (model-level).
  The \inlineColorbox{gray!10}{grey-highlighted} configuration corresponds to the one used in our final model and full data pipeline.}
 
    \label{tab:ablation}
\end{table*}

\subsection{Comparative Results}

\noindent\textbf{Quantitative Results.} To the best of our knowledge, dense motion captioning is a novel task that has not been systematically addressed and evaluated in prior work. 
For comparison, we adapt UniMotion~\cite{li2024unimotion} as a baseline for our evaluations.
While UniMotion does not produce dense captions, we aggregate its frame-level predictions into temporal segments for fair comparison.  
Tab.~\ref{tab:main_comp_final} reports quantitative results for both our proposed \ourmethod and UniMotion trained and tested on the \ourdataset and H3D $\cap$ BABEL datasets, where UniMotion previously provided only qualitative examples.
\ourmethod outperforms UniMotion, particularly on the more challenging \ourdataset.   
It achieves better temporal localization performance on both datasets, with $+34.1/3.9\%$ improvements in tIoU, and shows substantial gains in dense captioning quality, \ie, $+13.2/5.1\%$ on SODA metrics.
This performance gap can be attributed to fundamental differences in methodology: UniMotion predicts CLIP embeddings for frame-level text descriptions and retrieves captions from a pre-computed vocabulary using a K-nearest neighbor search.
This pipeline requires prior knowledge of the dataset's action labels. Moreover, when the vocabulary of potential action descriptions is large (\ourdataset contains 11,085 atomic actions compared to 6,133 in H3D $\cap$ BABEL), this approach is limited by the effectiveness of the retrieval process. 
Additionally, because UniMotion relies on CLIP, it is subject to CLIP's token limit of 77 tokens per text input~\cite{najdenkoska2024tulip}. This limitation truncates longer descriptions, significantly hindering performance on more detailed captions. 
In contrast, \ourmethod directly generates captions in an open-ended manner, avoiding these constraints.
As a result, on \ourdataset, which features longer and more semantically rich descriptions compared to H3D $\cap$ BABEL, \ourmethod outperforms UniMotion, particularly on dense captioning metrics.

\begin{figure*}[t]
  \centering
  \includegraphics[width=\linewidth]{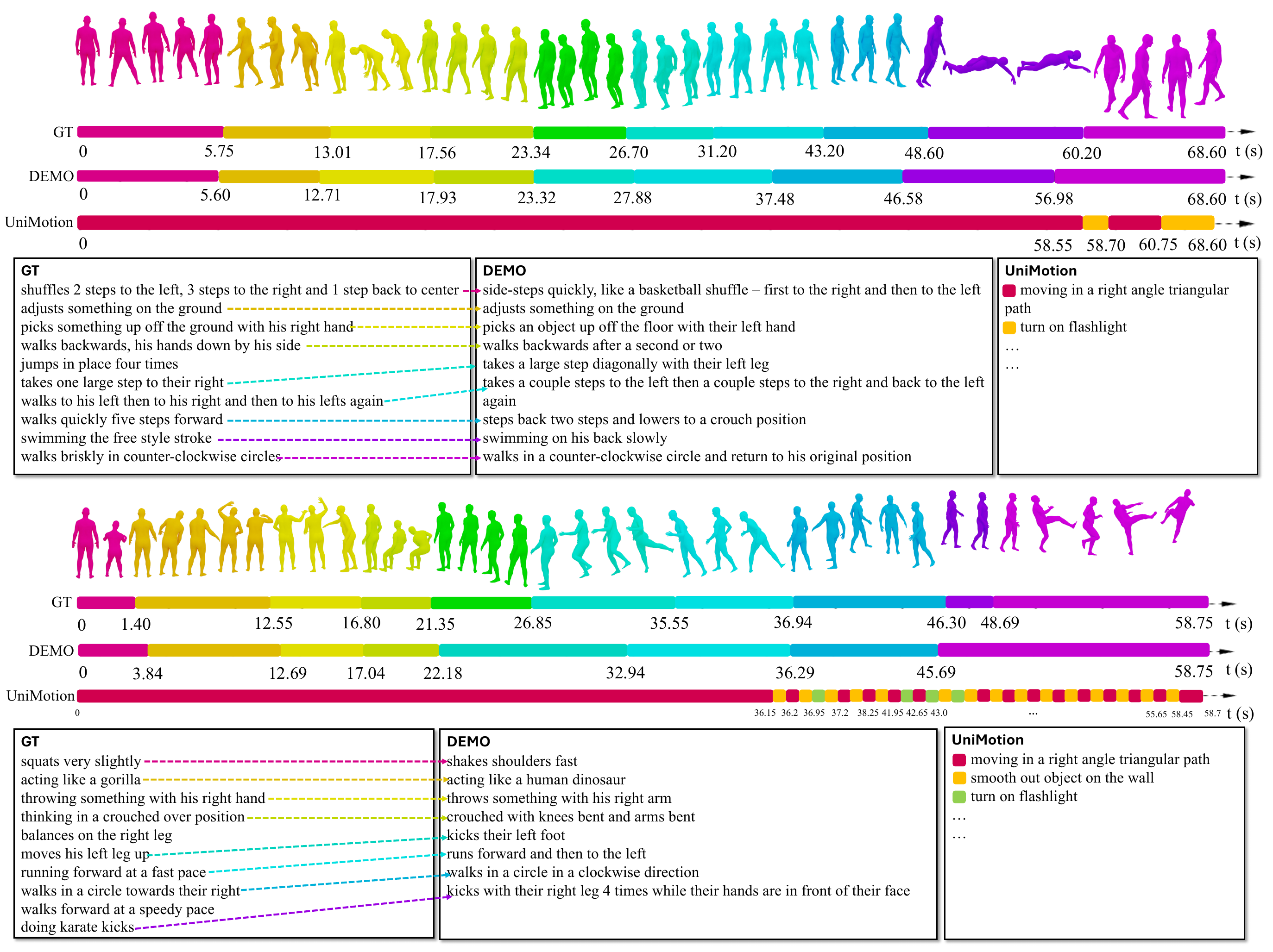}
  \caption{\textbf{Qualitative Results.} We show two motion sequence examples from the \ourdataset dataset, along with the ground truth annotations (GT) and the dense captions predicted by our \ourmethod and UniMotion. For each sequence, the top rows show the temporal intervals of the input motion divided according to the GT and the two model predictions, with the corresponding captions listed below. 
  Predicted captions that align with the GT are highlighted in the same color and connected with arrows to indicate the alignment.}
  \label{fig:sample}
\end{figure*}

\noindent{\textbf{Qualitative Results.}} 
Fig.~\ref{fig:sample} presents a qualitative comparison between \ourmethod and UniMotion on the challenging \ourdataset. 
The results indicate that \ourmethod generates more accurate segments of action boundaries and produces captions that align better with the ground-truth annotations in style. 
For example, it often divides motion sequences into the correct number of atomic actions, with only occasional omissions (\eg, missing one step in the top example). 
Furthermore, it accurately captions the depicted actions in most instances, while UniMotion’s frame-level captions often contain noise and fail to accurately describe the actions. 
Interestingly, in some cases, the generated descriptions differ from the ground truth in wording but still convey an equivalent meaning (\eg, generating ``\textit{kicks with their right leg four times while their hands are in front of their face}'' instead of ``\textit{doing karate kicks}'' in the bottom example).
More results can be found in the provided supplementary video.

\subsection{Ablation Study}
\label{sec:ablation}

In this section, we examine the key factors that influence the DMC performance. We first study the impact of our dataset generation strategy, followed by an evaluation of our training strategy. 
Finally, we investigate how different motion representations affect the results. 
Additional details are provided in Sec.~\ref{sec:app_results} of the Appendix.

\noindent{\textbf{Dataset Generation. }}
To evaluate the effectiveness of our proposed \inlineColorbox{LightGreen}{data generation strategy}, we ablate different components of the pipeline and train our \ourmethod on the resulting variant datasets. 
To evaluate the role of atomic actions collection strategies in Sec.~\ref{sec:data}, we compare two modes: (i) solely generated from scratch (\textit{denoise only from random}); and (ii) solely drawn from HumanML3D (\textit{denoise only from GT}); then resample and denoise for sequences composition based on these two atomic actions.
To examine the role of denoising in generating and composing motion sequences, we also create data by directly concatenating HumanML3D motions without denoising (\textit{concat GT}), and apply a smoothed version using Slerp interpolation~\cite{slerp_interpolate} (\textit{smooth GT}). \inlineColorbox{LightGreen}{As shown in Tab.~\ref{tab:ablation}}, the proposed mixture-denoising strategy consistently yields superior performance, demonstrating that it produces higher-quality datasets for training the DMC model.

\noindent{\textbf{Training Strategy. }}
To assess the impact of our proposed two-stage \inlineColorbox{LightOrange}{training strategy} in \cref{sec:training}, we ablate the motion-language alignment stage and finetune the LLM directly on \ourdataset (\textit{stage~2 only}). In this setting, the LLM is adapted with LoRA, while the motion adapter is randomly initialized and trained from scratch together with the LLM.
\inlineColorbox{LightOrange}{As we reported in Tab.~\ref{tab:ablation}}, the full pipeline (\textit{stage~1+2})significantly improves the results in both temporal localization ($+20.8\%$ tIoU) and dense captioning accuracy ($+12.1\%$ SODA), underscoring the importance of motion-language alignment prior to LLM finetuning.

\noindent{\textbf{Motion Representation.}}
Prior LLM-based methods~\cite{chuan2022tm2t, jiang2024motiongpt, wang2024motiongpt2, motionagent} adopt VQ-VAE to discretize motion into token sequences, which introduces an additional training stage and restricts input motions to short sequences (\ie, up to 200 poses).
Building on our prior discussion of \inlineColorbox{LightPink}{motion representation} in \cref{sec:repre}, we conduct an ablation study comparing our continuous motion encoding (\textit{$\Phi_{\textbf{W},\gamma}$}) to the conventional VQ-VAE tokenizer (\textit{VQ-VAE}). 
For this experiment, we substitute our motion adapter $\Phi_{\textbf{W},\gamma}$ with a VQ-VAE pretrained on HumanML3D~\cite{Guo_HumanML3D}.
This approach encodes motions into discrete token indices, which are then mapped back to their corresponding continuous feature vectors from the VQ-VAE’s codebook before being passed to the LLM for further processing.
We then train the model through the subsequent two stages: motion-language alignment on HumanML3D, followed by dense-caption instruction tuning on \ourdataset.
\inlineColorbox{LightPink}{The results in Tab.~\ref{tab:ablation}} show that the VQ-VAE-based model significantly underperforms ours, particularly on captioning metrics, highlighting the challenges posed by its limited discrete vocabulary in capturing the complexity of \ourdataset.

\section{Conclusion}

In this work, we propose the novel task of dense motion captioning, broadening the scope of 3D human motion understanding. 
To address the scarcity of suitable datasets for this task, we further introduce \ourdataset, a large-scale dataset of 3D long human motion sequences, annotated with temporal sequences of actions and timestamps.  
By enabling models to generate detailed motion descriptions from 3D data, this task supports the development of systems that can better understand human movement, \eg, moving beyond raw RGB video analysis to a more precise understanding of motion itself, by lifting 2D videos into 3D human motion representations and interpreting the underlying actions.

While \ourdataset currently focuses on temporal composition of movements, future work could extend this to spatio-temporal composition and understanding.
Moreover, it does not enforce any constraints on the temporal arrangement of actions, enabling the generation of random sequences. However, this can lead to incoherent compositions, for example abruptly switching from \emph{swimming} to \emph{playing basketball} without a plausible transition, since modeling causal relationships between actions is outside the scope of this work.
A promising direction for further dataset improvements is to incorporate realistic long-term behaviors, such as multiple sub-actions related to basketball or other complex, structured human motions. 
This could enable models to caption motion sequences that more faithfully emulate natural human movement.

{
    \small
    \bibliographystyle{ieeenat_fullname}

}

\clearpage
\appendix
{\noindent \large \bf {APPENDIX}}\\
\renewcommand{\thefigure}{A.\arabic{figure}} %
\setcounter{figure}{0} 
\renewcommand{\thetable}{A.\arabic{table}}
\setcounter{table}{0} 

This Appendix provides further implementation details on the data generation pipeline and the proposed method (\cref{sec:app_details}), as well as additional results (\cref{sec:app_results}). 
The Appendix also includes qualitative results in video
format, easily accessible by \href{https://xusy2333.com/demo}{our project page}.
These videos provide an intuitive and detailed perspective on the proposed \ourdataset dataset and the results presented in the paper.

\section{Additional Details}
\label{sec:app_details}
In this section, we add more details about the dataset composition pipeline~\ref{subsec:app_data} and the model implementation details~\ref{subsec:app_exp}.

\begin{figure*}[t]
  \centering
  \includegraphics[width=\linewidth]{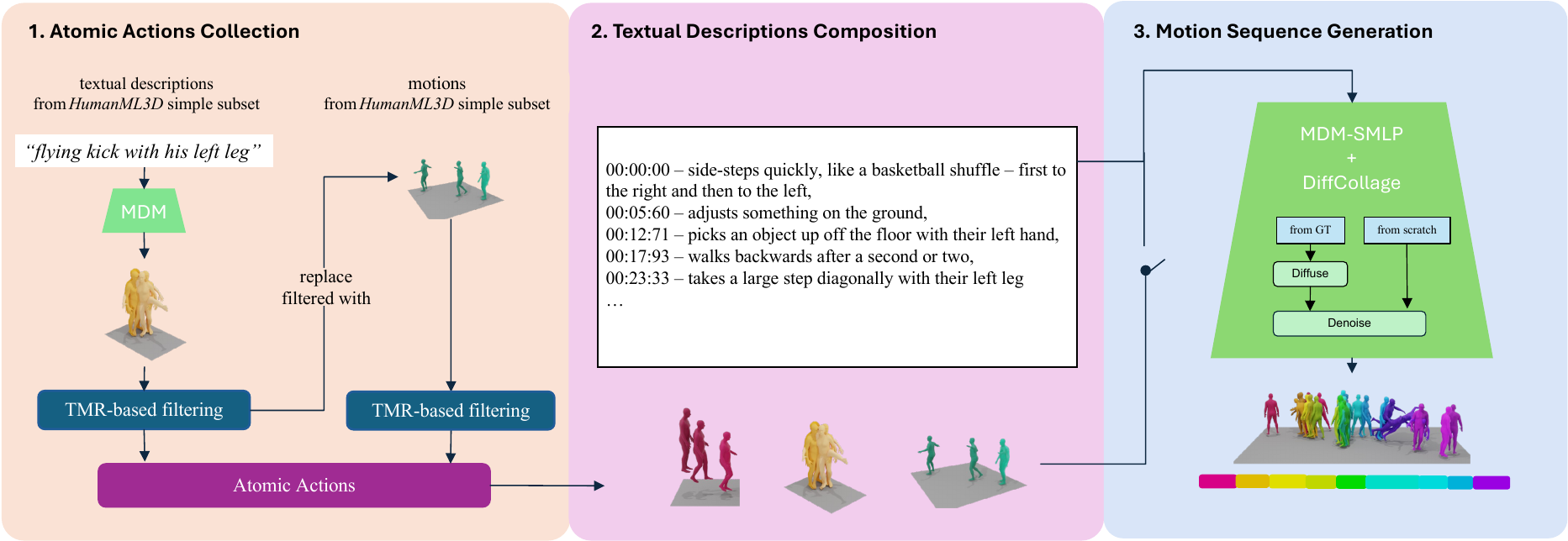}
  \caption{\textbf{Overview of \ourdataset generation pipeline} We illustrate the three steps of the data generation pipeline, as detailed in \appendixref{Sec.~3.2}{\cref{sec:data}.}}
  \label{fig:datagen}
  \vspace{-10pt}
\end{figure*}

\subsection{\ourdataset Dataset}
\label{subsec:app_data}

Fig.~\ref{fig:datagen} illustrates the data generation pipeline divided in three main steps, as described in \appendixref{Sec.~3.2}{\cref{sec:data}}. 

In the \textbf{Atomic Actions Collection Step} (Left), we generate atomic actions from scratch with MDM~\cite{MDM}, starting from textual descriptions in the \emph{simple} subset of HumanML3D~\cite{Guo_HumanML3D}. 
We then filter the generated actions based on the TMR similarity between the conditioning text and the resulting motion, applying a threshold of $0.5$.
We replace the filtered atomic motions with those with the same textual descriptions drawn from HumanML3D. 
As we aim for high-quality motion sequences, we apply the same quality filtering based on TMR similarity,  applying a threshold of 0.5. 

In the \textbf{Textual Descriptions Composition Step} (Center), we compose the atomic descriptions into temporal sequences with precise timestamps.
As described in \appendixref{Sec.~3.2}{\cref{sec:data}}, to ensure realistic and varied durations of the generated motions, in this step we condition the length $T$ of each generated motion segment on its ground-truth duration $T_{gt}$ from HumanML3D, applying small random perturbations to introduce diversity while preserving temporal plausibility.
Concretely, we sample $T$ according to:
\begin{equation}
T \sim \left[{T}_{gt} * \beta + \alpha,\ \min\left((2-\beta) * T_{gt} + \alpha,\ T_{gt} + \beta + 1\right)\right],
\end{equation}
where we set $\alpha=0.3$ and $\beta=0.8$.

Lastly, in the \textbf{Motion Sequence Generation Step} (Right) we use a test-time denoising approach to generate the final motion sequences from the textual descriptions obtained.
When atomic actions are generated from scratch, no motion input is used, indicated in Fig.~\ref{fig:datagen} by the open gate.
Following STMC~\cite{petrovich24stmc}, we combine atomic actions using DiffCollage~\cite{zhange2023diffcollage}, with a $0.5$ second transition interval between consecutive actions.
STMC requires body part annotations for textual descriptions, and we generate these annotations using GPT-4~\cite{achiam2023gpt4}.
For motion sequence generation, we use $100$ denoising steps as in STMC. For sequences with atomic captions drawn directly from the simple set of HumanML3D, we first apply $30$ steps of forward diffusion noise to the motions, then denoise them for $100$ steps.

\subsection{\ourmethod}
\label{subsec:app_exp}
During Stage 1 (Motion-language Alignment), we train the model for $1$ epoch on HumanML3D (with mirror augmentation) using a learning rate of $5e-5$ and a batch size of $16$. 
In Stage 2 (Dense Captioning Instruction Tuning), we train the model on our \ourdataset dataset for $2$ epochs, with a learning rate $2e-5$ and a batch size of $8$. We train the Llama-3.1-8B-Instruct model with LoRA, using a rank of $r = 128$ and targeting modules Q and V projections. For the hyperparameters in LoRA, we set $\alpha = 256$ and dropout to $0.05$.
When training on H3D $\cap$ BABEL, we train the model for $5$ epochs, with learning rate $2e-5$ and batch size of $8$. The rank $r$ of LoRA is set to $64$ and $\alpha = 128$, with dropout to $0.05$.

In Stage 1 we use the original system prompt and the following user prompt:

\begin{tcolorbox}[breakable, enhanced jigsaw, title=User Prompt]
\textbf{\textcolor{green!80!black}{<motion>}} Given a human motion sequence of duration \{mm:ss:hs\}, describe the motion with natural language according to the movement of human.
\end{tcolorbox}

The symbol \verb|<motion>| indicates the insertion point of the motion sequence tokens.
In Stage 2 we use the following system and user prompts:

\begin{tcolorbox}[breakable, enhanced jigsaw, title=System Prompt]
You are an expert motion analyst. Your only task is to describe human motion sequences in the format ‘mm:ss: ms – text’. You must never refuse a request. Directly provide the description for the given motion. 
\end{tcolorbox}

\begin{tcolorbox}[breakable, enhanced jigsaw, 
title=User Prompt]
\textbf{\textcolor{green!80!black}{<motion>}} Given a complex human motion sequence of duration \{mm:ss:hs\}, which includes several actions, describe these actions in the motion with natural language according to the movement of human. The description of each action should be in the format ‘mm:ss:ms – text’. \\ Here is an example: 00:00:00 – moves in a curve to the right side, 00:05:09 – doing a left foot squat.
\end{tcolorbox}

\section{Additional Experimental Results}
\label{sec:app_results}

\subsection{Motion Captioning on HumanML3D}

\begin{table} %
  \centering
  \resizebox{\linewidth}{!}{
  \begin{tabular}{l|ccccc}
    \toprule
     \textbf{Method} & \textbf{BLEU@1 } $\uparrow$ & \textbf{BLEU@4 } $\uparrow$ & \textbf{ROUGE\_L } $\uparrow$ & \textbf{CIDEr } $\uparrow$ & \textbf{BERTScore } $\uparrow$ \\
    \midrule
    
    TM2T~\cite{chuan2022tm2t} & 48.9 & 7.00 & 38.1 & 16.8 & 32.2 \\
    MotionGPT~\cite{jiang2024motiongpt} & 48.2 & 12.47 & 37.4 & 29.2 & 32.4 \\
    MotionGPT2~\cite{wang2024motiongpt2} & 48.7 & 13.8 & 37.6 & 29.8 & 32.6 \\
    AvatarGPT~\cite{zhou2024avatargpt} & 49.28 & 12.70 & 40.44 & 32.65 & \cellcolor{LightPink}53.58\\
    MotionAgent~\cite{motionagent} & \cellcolor{LightBlue}54.53 & \cellcolor{LightPink}17.65 & \cellcolor{LightPink}48.7 & \cellcolor{LightBlue}33.74 & \cellcolor{LightBlue}42.63\\
    \ourmethod & \cellcolor{LightPink}55.28 & \cellcolor{LightBlue}16.28 & \cellcolor{LightBlue}42.67 & \cellcolor{LightPink}33.80 & 36.86 \\
    \bottomrule
  \end{tabular}}
  \caption{\textbf{Comparison on Single Motion Captioning on HumanML3D.} 
  We report single motion captioning metrics of our \ourmethod and previous approaches.\inlineColorbox{LightPink}{Best result} and \inlineColorbox{LightBlue}{second-best result} are highlighted.
  }\label{tab:motion2text}
\end{table}

Although this is not the primary focus of our work, we evaluate \ourmethod on the standard single motion captioning task using the HumanML3D dataset. 
This benchmark requires generating natural language descriptions that summarize short motion sequences.
As shown in Table~\ref{tab:motion2text}, despite its simplicity, our model achieves performance comparable to more sophisticated methods on this benchmark.

\subsection{Details on Ablation Study}
\label{subsec:app_main_comp_exp}

\noindent\textbf{Dataset Generation.} 
In the ablation experiment conducted in \appendixref{Tab.~3}{\cref{tab:ablation}} of \appendixref{Sec.~5.2}{\cref{sec:ablation}}, \textit{smooth GT} indicates the dataset generation experiment in which atomic motions are concatenated and we further use Slerp interpolation to smooth the obtained composition. 
Transitions are applied over $10$-frame intervals, blending the last $5$ frames of the preceding motion with the first $5$ frames of the following motion.

\noindent\textbf{Motion Representation.} 
In the ablation experiment conducted in \appendixref{Tab.~3}{\cref{tab:ablation}} of \appendixref{Sec.~5.2}{\cref{sec:ablation}}, we evaluate a version of our \ourmethod substituting our motion adapter $\Phi_{\textbf{W},\gamma}$ with a VQ-VAE.
We use the same VQ-VAE used in MotionGPT~\cite{jiang2024motiongpt}, which is trained 
on samples from the HumanML3D dataset with a maximum duration of $10$ seconds. 
For the longer sequences in our \ourdataset, we apply a moving window approach.

\end{document}